\pdfoutput=1

\documentclass[11pt]{article}
\usepackage{acl2023}
\usepackage{times}
\usepackage{latexsym}
\usepackage[T1]{fontenc}
\usepackage[utf8]{inputenc}
\usepackage{microtype}
\usepackage{url}
\usepackage{graphicx}
\usepackage{tabularx}
\usepackage{xcolor}
\usepackage{booktabs}
\usepackage{xspace}
\usepackage{amsmath}
\usepackage{amsfonts}
\usepackage{xfp}
\usepackage{multirow}
\usepackage{subfig}
\usepackage{censor}
\usepackage{scalefnt}
\usepackage{cleveref}
\usepackage{inconsolata}
\usepackage{stackengine}
\usepackage{hyperref}

\setstackgap{L}{.7\normalbaselineskip}
\stackMath

\newcommand{\emotionname}[1]{\textit{#1}}
\newcommand{\fear}{\emotionname{fear}\xspace}
\newcommand{\joy}{\emotionname{joy}\xspace}
\newcommand{\anger}{\emotionname{anger}\xspace}

\newcommand{\guilt}{\emotionname{guilt}\xspace}
\newcommand{\shame}{\emotionname{shame}\xspace}

\newcommand{\sadness}{\emotionname{sadness}\xspace}

\newcommand{\disgust}{\emotionname{disgust}\xspace}

\newcommand{\other}{\emotionname{other}\xspace}
\newcommand{\noemotion}{\emotionname{no emotion}\xspace}

\newcommand{\effort}{\emotionname{effort}\xspace}

\newcommand{\outProb}{\emotionname{outcome probability}\xspace}
\newcommand{\goalConducive}{\emotionname{goal conduciveness}\xspace}
\newcommand{\adjCheck}{\emotionname{adjustment check}\xspace}

\newcommand{\otherResp}{\emotionname{other responsibility}\xspace}

\newcommand{\attention}{\emotionname{attention}\xspace}
\newcommand{\selfResp}{\emotionname{self responsibility}\xspace}
\newcommand{\responsibility}{\emotionname{responsibility}\xspace}

\newcommand{\internCheck}{\emotionname{internal check}\xspace}
\newcommand{\control}{\emotionname{control}\xspace}

\newcommand{\attend}{\emotionname{attend}\xspace}

\newcommand{\xenvcorpus}{x-en\textsc{Vent}\xspace}

\newcommand{\F}{F$_1$\xspace}

\renewcommand{\paragraph}[1]{\noindent\textbf{#1}}

\title{Automatic Emotion Experiencer Recognition}

\author{%
  Maximilian Wegge \and Roman Klinger \\
  Institut f\"ur Maschinelle Sprachverarbeitung, University of Stuttgart \\
  \texttt{\{firstname.lastname\}@ims.uni-stuttgart.de}
}

\begin{document}
\maketitle
\begin{abstract}
  The most prominent subtask in emotion analysis is emotion
  classification; to assign a category to a textual unit, for instance
  a social media post. Many research questions from the social
  sciences do, however, not only require the detection of the emotion
  of an author of a post but to understand who is ascribed an emotion
  in text. This task is tackled by emotion role labeling which aims at
  extracting who is described in text to experience an emotion, why,
  and towards whom. This could, however, be considered overly
  sophisticated if the main question to answer is who feels which
  emotion. A targeted approach for such setup is to classify emotion
  experiencer mentions (aka ``emoters'') regarding the emotion they
  presumably perceive. This task is similar to named entity
  recognition of person names with the difference that not every
  mentioned entity name is an emoter. While, very recently, data with
  emoter annotations has been made available, no experiments have yet
  been performed to detect such mentions. With this paper, we provide
  baseline experiments to understand how challenging the task is. We
  further evaluate the impact on experiencer-specific emotion
  categorization and appraisal detection in a pipeline, when gold
  mentions are not available.  We show that experiencer detection in
  text is a challenging task, with a precision of .82 and a recall of
  .56 (\F=.66). These results motivate future work of jointly modeling
  emoter spans and emotion/appraisal predictions.
\end{abstract}
\section{Introduction}
\label{sec:intro}
Computational emotion classification is among the most prominent tasks
in the field of textual emotion analysis. It is typically
formulated as either a classification or regression task, depending on
the underlying emotion theory and intended application and domain:
Texts can be classified into one or multiple discrete emotion
categories, following the concept of basic emotions by
\citet{Ekman1992} or \citet{Plutchik2001}, as continuous values within
the vector space of valence, arousal and dominance
\citep{russell1977evidence} or based on the emoter's cognitive
appraisal of the emotion-eliciting event (e.g., the level of \control
or \responsibility; \citealp{Smith1985}).
	
Recent work has emphasized the relevance of perspective, i.e., whose
emotion is considered given an emotion-eliciting event.  Typically,
emotions are investigated from either the writer's or the reader's
perspective, with only few approaches that consider both
\citep[e.g.,][]{buechel-hahn-2017-readers}.
Although not exclusively focused on it, perspective is also addressed
in the context of semantic role labeling (``Who is feeling the
emotion?''), besides the emotion target (``Who is the emotion directed
towards?'') and cause (``What is causing the emotion?'')
\citep{mohammad-etal-2014-semantic,Bostan2020-gne}.
\citet{Troiano2022-xenvent} build upon this idea and extend the
investigation to all potential emoters affected by an event.  For each
entity, they consider their emotions and the appraisal of the
corresponding event, which allows to disambiguate the individual
emotions.

Consider the example ``Ken Paxton: Texas House votes to impeach Trump
ally''\footnote{\scalebox{0.91}[1]{\url{https://www.bbc.com/news/world-us-canada-65736478}}}. Here,
``Ken Paxton'' could be attributed \guilt because of the impeachment
process following a potential appraisal of \selfResp. ``Trump'' being
described as an ally might develop \anger because he might evaluate
the situation differently and assign an appraisal of
\otherResp. ``Texas House'' could be considered a named entity, but
does not represent an emoter. The writer's emotion is presumably
irrelevant in such news headline. Experiencer-agnostic approaches can
only assign emotions and appraisal to the entire text, thus
oversimplifying the relations between individual experiencers.

\citet{wegge-etal-2022-experiencer} compare experiencer- and
text-level emotion/appraisal predictors on self-reported event
descriptions.  They find that an experiencer-specific predictor is
able to capture the individual information, while a conventional
classifier averages over all individual (potentially contradictory)
information in the entire text.  While they provide a computational
approach for experiencer-specific emotion and appraisal
classification, they rely on gold annotations of experiencer-spans.
They do not investigate whether these spans can be predicted reliably
and what consequences this would have on the classification task.
	
In this paper, we evaluate (i.) the performance of an automatic
experiencer-detection model and (ii.) the impact of the imperfect
automatic prediction on emotion and
appraisal classification.  We show that there is a substantial drop in
the pipeline model in contrast to using gold annotations, which
motivates future joint modeling work.

\section{Related Work}
\label{sec:relatedwork}
Computational emotion classification is commonly grounded in theories
of basic emotions, i.e., \citet{Ekman1992} or \citet{Plutchik2001},
while regression models often handle emotions as tuples of continuous
values within a vector space, for instance of valence, arousal, and
dominance \citep{russell1977evidence}. Emotion intensity prediction
combines both classification and regression tasks by assigning not
only an emotion category but a corresponding intensity score as well
\citep{mohammad-bravo-marquez-2017-emotion}.  In appraisal theories,
emotions depend on the emoter's cognitive evaluation of the event
\citep{Smith1985,Scherer2001a} and are either defined by it directly
or are understood to emerge out of it, depending on the respective
theory \citep{Scarantino2016}.

This cognitive appraisal can be modeled with variables that represent
the emoter's event evaluation, for instance whether the emoter could
anticipate the consequences of the event (\outProb) or whether the
emoter is responsible for what is happening (\selfResp) rather than
another entity (\otherResp). The appraisal theories make an obvious
aspect explicit: the emotion is developed by an entity that is part of
an emotional episode. This work therefore puts emphasis not only on a
cause or expression of an emotion, but also by whom it is perceived.

Emotion classification received substantial attention in a variety of
domains like social media posts
(\citealp{mohammad-bravo-marquez-2017-emotion};
\citealp{stranisci-etal-2022-appreddit}; i.a.), news headlines
\citep{Bostan2020-gne} or literary texts \citep{Ovesdotter2005}.  Most
work focused on the emotions from a single perspective.  Semantic role
labeling does consider more than one perspective, but is primarily
focused on the relations between experiencers, targets, and causes
(\citealp{Bostan2020-gne}; \citealp{mohammad-etal-2014-semantic};
\citealp{Kim2018}). The work on emotion experiencer detection is a
more direct access to the emotion experiencer
\citep{wegge-etal-2022-experiencer,Troiano2022-xenvent}. In comparison
to emotion role labeling, that is a simplification that enables a more
straight-forward modeling. These modeling differences are similar to
representing aspect-based sentiment analysis as an aspect
classification task rather than finding full graph representations of
evaluative phrases and mentioned aspects \citep[compare the two shared
task setups described
by][]{barnes-etal-2022-semeval,pontiki-etal-2014-semeval}.

Appraisal theories already motivated some NLP research
\citep{Troiano2023,Hofmann2020b,stranisci-etal-2022-appreddit}, but
only recently, \citet{Troiano2022-xenvent} investigate all potential
perspectives involved in an event with their \xenvcorpus corpus, based
on self-reported event descriptions \citep{Troiano2019-enisear}.  The
corpus is annotated with potential emoters, their respective emotions
and 22 appraisals (score from 0--5 for each
dimension). \citet{wegge-etal-2022-experiencer} proposed first models
to assign emotions and appraisals to experiencer mentions, but did
rely on the experiencer annotations. Therefore, it is still an open
research question what the challenges of emotion experiencer detection
are; the gap that we aim at filling with this paper.
 
\section{Methods}
\label{sec:methods}
Our methods consists of a pipeline of (a) experiencer detection followed
by (b) experiencer-aware emotion/appraisal detection. For the second
step, we follow \citet{wegge-etal-2022-experiencer} who purely relied
on gold annotations for the first step.

The experiencers consist of sequences of tokens within a text (we
assume experiencer-spans to be non-overlapping). The writer's
perspective is represented with such annotation on a special token
prefix \verb|writer|.  One text can contain multiple experiencer
spans.  Each experiencer gets assigned a set of emotion labels (6
Ekman emotions + other, no emotion, and shame) and a set of up to 22
appraisal dimensions (see Table~\ref{tab:apprresults} for a list of
classes).

Our pipeline consists of two steps: (i.) the detection of
experiencers and (ii.) the prediction of emotions/appraisal
dimensions for each experiencer.

\paragraph{Models.}
For detecting the experiencer-spans, we fine-tune a transition-based
named entity recognition model (NER) from the spaCy library
\citep{honnibal-etal-spacy} on the \xenvcorpus corpus
\citep{Troiano2022-xenvent}.  The data set consists of 720 instances
which we split into 538 for training (of which we use 61 for
validation) and 107 for testing. We omit 14 instances that contain
overlapping spans.\footnote{We use the default spaCy configuration,
  learning rate 0.001, weight decay, dropout 0.1, Adam optimizer.}

Our goal is to ensure comparability with previous work on
experiencer-specific emotion and appraisal classification. Therefore,
we apply the same models as \citet{wegge-etal-2022-experiencer}, by
fine-tuning Distil-RoBERTa (\citealp{Liu2019Roberta}, using Hugging
Face's transformers library, \citealp{Wolf2020}) with a multi-output
classification head to jointly predict all emotion labels (see their
paper for implementation details).  Experiencer-spans are encoded via
positional indicators in the text \citep[cf.][]{Zhou2016}.  We differ
from the previous approach in formulating the prediction of appraisal
dimensions as classification instead of regression to have a
straight-forward access to an evaluation of the overall pipeline in
which additional experiencers might appear that are not available in
the gold annotation.  To this end, we use a threshold of $4$ to
discretize the continuous appraisal scores.  The appraisal
classification head is analogous to the one for emotions.\footnote{Our code is available
  at \url{https://www.ims.uni-stuttgart.de/data/appraisalemotion}.}

\paragraph{Evaluation.} We evaluate the performance of our pipeline by
calculating the \F in two settings.  In the \textit{strict}
evaluation, only exact matches of token spans make true positives. In
the \textit{relaxed} setting, we additionally accept partial matches
with at least one token overlap as true positives.

We apply the experiencer-specific classifiers to the experiencer-spans
detected in the first pipeline component instead of the gold spans.
We consider this in the calculation of \F by treating every predicted
emotion or appraisal label as a false positive if the associated
experiencer-span has no correspondence in the gold data (we accept
overlapping spans).  Analogously, if a gold experiencer-span was not
recognized by the experiencer-span detector, we consider each gold
emotion and appraisal label that was associated with that span a false
negative.  We compare our results against the performance values on
gold-annotated experiencer spans.

\section{Results}
\label{sec:results}
\begin{table}[t]
  \centering\small
  \setlength{\tabcolsep}{6pt}
  \begin{tabular}{l cc cc cc}
    \toprule
   & \multicolumn{2}{c}{P} & \multicolumn{2}{c}{R} &\multicolumn{2}{c}{\F} \\
    \cmidrule(rl){2-3}\cmidrule(rl){4-5}\cmidrule(l){6-7}
    & s & r &s &r&s&r\\
    \cmidrule(rl){2-2}\cmidrule(rl){3-3}\cmidrule(rl){4-4}\cmidrule(rl){5-5}\cmidrule(lr){6-6}\cmidrule(l){7-7}
    incl.\ \textsc{writer}      & 90&93 & 77&80 & 83&86\\
    excl.\ \textsc{writer}      & 74&82& 50&56 & 60&66\\
    \bottomrule
  \end{tabular}
  \caption{Span-prediction results (s: strict; r: relaxed).}
  \label{tab:res-spans}
\end{table}
We report results for both pipeline components.
\subsection{Experiencer-Span Detection}
\label{subsec:res-spans}
\autoref{tab:res-spans} reports the
precision, recall and \F of the span-detector for all non-writer
experiencers (excl.\ \textsc{writer}) as well as to all
experiencer-spans (incl.\ \textsc{writer}). Recognizing the
$\verb|writer|$ token as an experiencer is trivial (\F=1.0).

As to be expected, the performance of the span-predictor is lower in
the evaluation setup that considers only the non-writer experiencers.
There is a considerable difference in the exact and relaxed evaluation
setup, which shows that the model sometimes only finds a subset of the
experiencer tokens. The task is challenging: while the precision is
acceptable, only half of the experiencers are found. This is to some
degree a result of the annotation of the data -- the corpus authors
tasked the annotators to only label the first occurrence of each
mention of an experiencer in a text -- a property that is challenging
to be grasped automatically.
\subsection{Emotion and Appraisal Classification}
\label{subsec:res-class}
\begin{table}
  \centering\small
  \setlength{\tabcolsep}{6pt}
  \begin{tabular}{l rrr rrr r}
    \toprule
    & \multicolumn{3}{c}{\textsc{gold spans}} & \multicolumn{3}{c}{\textsc{pipeline}}&\\
    \cmidrule(r){2-4}
    \cmidrule(r){5-7}
    Emotion 
    & P & R & \F & P & R & \F & $\Delta$\F \\
    \cmidrule(r){1-1}\cmidrule(rl){2-4}\cmidrule(rl){5-7}\cmidrule(l){8-8}
    anger      & 73 & 53 & 61 & 77 & 45 & 57 & $-4$ \\
    disgust     & 76 & 81 & 79 & 64 & 56 & 60 & $-19$  \\
    fear       & 82 & 60 & 69 & 68 & 57 & 62 & $-7$  \\
    joy        & 48 & 82 & 60 & 49 & 69 & 57 & $-3$ \\
    no emotion & 54 & 79 & 64 & 47 & 47 & 47 & $-17$ \\
    other      & 33 & 5 & 9 & 50 & 5 & 9  & $\pm0$ \\
    sadness    & 61 & 77 & 68 & 57 & 65 & 61 & $-7$ \\
    shame      & 57 & 73 & 64 & 54 & 59 & 56 & $-8$ \\
    
    \cmidrule(r){1-1}\cmidrule(rl){2-4}\cmidrule(rl){5-7}\cmidrule(l){8-8}
    Macro avg. & 49 & 66 & 56 & 40 & 62 & 49 & $-7$  \\
    Micro avg. & 55 & 72 & 62 & 43 & 67 & 52 & $-10$ \\
    \bottomrule
  \end{tabular}
  \caption{The experiencer-specific emotion classifier is evaluated on expert-annotated (\textsc{gold spans}) and automatically detected (\textsc{pipeline}) experiencer-spans.}
  \label{tab:emoresults}
\end{table}
\autoref{tab:emoresults} reports the results of the emotion classifier
applied to the automatically predicted experiencer-spans
(\textsc{pipeline} setting) as well as the baseline results
(\textsc{gold spans}) that were obtained on expert-annotated
experiencer-spans.  Across almost all emotion categories, the
\textsc{pipeline} classifier performs worse than the \textsc{gold
  spans} baseline, which is expected as the evaluation method
penalizes erroneously detected experiencer-spans.  However, the drop
in performance differs between emotions. For \anger, \joy, \sadness,
\fear, \shame the difference is less than 10pp \F -- for these
emotions, experiencers can be found more reliably than for \disgust
(19pp) or \noemotion (17pp).

The notable decrease in performance for \noemotion is in line with the
observation that predicting non-writer spans is more challenging than
predicting writer-spans.  From all spans annotated with \noemotion,
84\% are non-writer spans.  However, the classification performance
also drops for emotion classes that are frequently annotated in
writer-spans; The pipeline classifier shows its biggest decrease in
performance (19pp) for \disgust, although 76\% of all spans annotated
with \disgust are writer-spans.  This is due to the span-predictor's
low recall: a low number of recognized spans leads to a higher number
of false negatives for all emotion classes associated with these
spans. The biggest increase in FN introduced by the span-predictor is
observed for \disgust (71\%), the lowest for \other (21\%).

\begin{table}
  \centering\small
  \setlength{\tabcolsep}{4.7pt}
  \begin{tabular}{l rrr rrr r}
    \toprule
    & \multicolumn{3}{c}{\textsc{gold spans}} & \multicolumn{3}{c}{\textsc{pipeline}}&\\
    \cmidrule(r){2-4}
    \cmidrule(r){5-7}
    Appraisal 
    & P & R & \F & P & R & \F & $\Delta$\F \\
    \cmidrule(r){1-1}\cmidrule(rl){2-4}\cmidrule(rl){5-7}\cmidrule(l){8-8}
    suddenness 			& 67 & 65 & 66 & 64 & 59 & 62 & $-3$ \\
    familiarity				 & 0 & 0 & 0 & 0 & 0 & 0 & $\pm0$ \\
    pleasantness		& 8 & 87 & 83 & 78 & 78 & 78 & $-9$ \\
    understand           & 80 & 100 & 89 & 77 & 82 & 80 & $-9$ \\
    goal relev.				& 38 & 33 & 0.35 & 29 & 22 & 25 & $-10$ \\
    self resp.      		 & 64 & 95 & 76 & 61 & 70 & 65 & $-11$ \\
    other resp.    		   & 73 & 73 & 73 & 64 & 60 & 62 & $-11$ \\
    sit. resp.                & 52 & 79 & 62 & 45 & 68 & 54 & $-8$ \\
    effort 					   & 67 & 29 & 40 & 20 & 14 & 17 & $-23$ \\
    exert   				   & 0 & 0 & 0 & 0 & 0 & 0  & $\pm0$ \\
    attend   				 & 50 & 17 & 25 & 50 & 17 & 25 & $\pm0$ \\
    consider   		         & 72 & 66 & 69 & 65 & 57 & 61 & $-8$ \\
    outcome prob.      & 55 & 75 & 63 & 51 & 62 & 56 & $-7$ \\
    expect. discrep.    & 72 & 63 & 67 & 67 & 56 & 61 & $-6$ \\
    goal conduc.         & 59 & 62 & 60 & 60 & 57 & 59 & $-1$ \\
    urgency  			   & 0 & 0 & 0 & 0 & 0 & 0 & $\pm0$ \\
    self control          & 58 & 89 & 70 & 58 & 64 & 61 & $-9$ \\
    other control        & 75 & 55 & 63 & 63 & 45 & 52 & $-11$ \\
    sit. control            & 52 & 78 & 62 & 46 & 67 & 55 & $-7$ \\
    adj. check            & 75 & 75 & 75 & 72 & 53 & 61 & $-14$ \\
    int. check             & 33 & 12 & 18 & 25 & 12 & 17 & $-1$ \\
    ext. check           & 0 & 0 & 0  & 0 & 0 & 0 & $\pm0$ \\

    \cmidrule(r){1-1}\cmidrule(rl){2-4}\cmidrule(rl){5-7}\cmidrule(l){8-8}
    Macro avg. & 46 & 64 & 54 & 42 & 48 & 45 & $-9$  \\
    Micro avg. & 58 & 86 & 69 & 54 & 69 & 61 & $-8$ \\
    \bottomrule
  \end{tabular}
  \caption{Appraisal classification results of the appraisal classifier evaluated on expert-annotated (\textsc{gold spans}) and automatically detected (\textsc{pipeline}) experiencer-spans.}
  \label{tab:apprresults}
\end{table}
 
Analogous to the emotion classifier, we observe a decrease in
performance for the appraisal predictor, reported in
\autoref{tab:apprresults}. Again, there is a substantial difference in
the drop of performance, with \effort and \adjCheck showing the
highest loss (23pp and 14pp, respectively) and \goalConducive,
\internCheck, \attend being the lowest (1pp or no difference).  Both
\effort and \adjCheck appear only seldom in writer-spans (33\% each),
while \goalConducive, \internCheck and \attend appear more often in
writer spans (between 39\% and 44\%) and are less prone to
unrecognized spans (44\%/40\% of FN are introduced through missing
spans for \goalConducive/\attention, 29\% for \internCheck; cf. \autoref{tab:appr_fn}).
However, the individual differences are less pronounced than for the
emotion classification results, due to the sparseness of some
appraisal dimensions.

We show more detailed emotion/appraisal-specific statistics of writer
spans and false negatives in the appendix.

\section{Discussion and Conclusion}
In this paper, we presented the first evaluation of experiencer
detection in text and the impact of these predictions on the
emotion/appraisal classification. We found that experiencer detection
is challenging but the results are promising.

The emotion/appraisal detection interacts with the span prediction
task.  This indicates that a joint model that can explore interactions
between experiencer and emotion/appraisal dimensions might work better
than the pipeline setting. Such model is however not trivial to be
build, because the emotion/appraisal classification depends on a
variable number of spans. Possible approaches include a purely
token-level classification task or multiple sequence labeling
setups. Such engineering attempts can also find inspiration in
emotion--cause pair extraction models
\citep[e.g.,][]{yuan-etal-2020-emotion}.

Our work also motivates other follow-up studies, namely to extend the
experiments to corpora that are fully annotated with emotion role
graphs \citep{campagnano-etal-2022-srl4e}, from which some contain
experiencer annotations
\citep{bostan-etal-2020-goodnewseveryone,kim-klinger-2018-feels,mohammad-etal-2014-semantic}. We
expect our approach to show improvements over full graph predictions
for the subtask of experiencer-specific emotion prediction due to
fewer model parameters.

\section*{Acknowledgements}

This research is funded by the German Research Council (DFG), project
``Computational Event Analysis based on Appraisal Theories for Emotion
Analysis''
CEAT (project number KL 2869/1-2).
\bibliographystyle{acl_natbib}
\bibliography{lit}

\begin{thebibliography}{28}
\expandafter\ifx\csname natexlab\endcsname\relax\def\natexlab#1{#1}\fi

\bibitem[{Alm et~al.(2005)Alm, Roth, and Sproat}]{Ovesdotter2005}
Cecilia~Ovesdotter Alm, Dan Roth, and Richard Sproat. 2005.
\newblock \href {https://www.aclweb.org/anthology/H05-1073} {Emotions from
  text: Machine learning for text-based emotion prediction}.
\newblock In \emph{Proceedings of Human Language Technology Conference and
  Conference on Empirical Methods in Natural Language Processing}, pages
  579--586, Vancouver, British Columbia, Canada. Association for Computational
  Linguistics.

\bibitem[{Barnes et~al.(2022)Barnes, Oberlaender, Troiano, Kutuzov, Buchmann,
  Agerri, {\O}vrelid, and Velldal}]{barnes-etal-2022-semeval}
Jeremy Barnes, Laura Oberlaender, Enrica Troiano, Andrey Kutuzov, Jan Buchmann,
  Rodrigo Agerri, Lilja {\O}vrelid, and Erik Velldal. 2022.
\newblock \href {https://doi.org/10.18653/v1/2022.semeval-1.180} {{S}em{E}val
  2022 task 10: Structured sentiment analysis}.
\newblock In \emph{Proceedings of the 16th International Workshop on Semantic
  Evaluation (SemEval-2022)}, pages 1280--1295, Seattle, United States.
  Association for Computational Linguistics.

\bibitem[{Bostan et~al.(2020{\natexlab{a}})Bostan, Kim, and
  Klinger}]{Bostan2020-gne}
Laura Ana~Maria Bostan, Evgeny Kim, and Roman Klinger. 2020{\natexlab{a}}.
\newblock \href {https://aclanthology.org/2020.lrec-1.194}
  {{G}ood{N}ews{E}veryone: A corpus of news headlines annotated with emotions,
  semantic roles, and reader perception}.
\newblock In \emph{Proceedings of the 12th Language Resources and Evaluation
  Conference}, pages 1554--1566, Marseille, France. European Language Resources
  Association.

\bibitem[{Bostan et~al.(2020{\natexlab{b}})Bostan, Kim, and
  Klinger}]{bostan-etal-2020-goodnewseveryone}
Laura Ana~Maria Bostan, Evgeny Kim, and Roman Klinger. 2020{\natexlab{b}}.
\newblock \href {https://aclanthology.org/2020.lrec-1.194}
  {{G}ood{N}ews{E}veryone: A corpus of news headlines annotated with emotions,
  semantic roles, and reader perception}.
\newblock In \emph{Proceedings of the Twelfth Language Resources and Evaluation
  Conference}, pages 1554--1566, Marseille, France. European Language Resources
  Association.

\bibitem[{Buechel and Hahn(2017)}]{buechel-hahn-2017-readers}
Sven Buechel and Udo Hahn. 2017.
\newblock \href {https://doi.org/10.18653/v1/W17-0801} {Readers vs. writers vs.
  texts: Coping with different perspectives of text understanding in emotion
  annotation}.
\newblock In \emph{Proceedings of the 11th Linguistic Annotation Workshop},
  pages 1--12, Valencia, Spain. Association for Computational Linguistics.

\bibitem[{Campagnano et~al.(2022)Campagnano, Conia, and
  Navigli}]{campagnano-etal-2022-srl4e}
Cesare Campagnano, Simone Conia, and Roberto Navigli. 2022.
\newblock \href {https://doi.org/10.18653/v1/2022.acl-long.314} {{SRL4E} {--}
  {S}emantic {R}ole {L}abeling for {E}motions: {A} unified evaluation
  framework}.
\newblock In \emph{Proceedings of the 60th Annual Meeting of the Association
  for Computational Linguistics (Volume 1: Long Papers)}, pages 4586--4601,
  Dublin, Ireland. Association for Computational Linguistics.

\bibitem[{Ekman(1992)}]{Ekman1992}
Paul Ekman. 1992.
\newblock \href {https://doi.org/10.1080/02699939208411068} {An argument for
  basic emotions}.
\newblock \emph{Cognition \& emotion}, 6(3-4):169--200.

\bibitem[{Hofmann et~al.(2020)Hofmann, Troiano, Sassenberg, and
  Klinger}]{Hofmann2020b}
Jan Hofmann, Enrica Troiano, Kai Sassenberg, and Roman Klinger. 2020.
\newblock \href {https://doi.org/10.18653/v1/2020.coling-main.11} {Appraisal
  theories for emotion classification in text}.
\newblock In \emph{Proceedings of the 28th International Conference on
  Computational Linguistics}, pages 125--138, Barcelona, Spain (Online).
  International Committee on Computational Linguistics.

\bibitem[{Honnibal et~al.(2020)Honnibal, Montani, Van~Landeghem, and
  Boyd}]{honnibal-etal-spacy}
Matthew Honnibal, Ines Montani, Sofie Van~Landeghem, and Adriane Boyd. 2020.
\newblock \href {https://doi.org/10.5281/zenodo.1212303} {{spaCy:
  Industrial-strength Natural Language Processing in Python}}.

\bibitem[{Kim and Klinger(2018{\natexlab{a}})}]{Kim2018}
Evgeny Kim and Roman Klinger. 2018{\natexlab{a}}.
\newblock \href {http://aclweb.org/anthology/C18-1114} {Who feels what and why?
  annotation of a literature corpus with semantic roles of emotions}.
\newblock In \emph{Proceedings of the 27th International Conference on
  Computational Linguistics}, pages 1345--1359, Santa Fe, New Mexico, USA.
  Association for Computational Linguistics.

\bibitem[{Kim and Klinger(2018{\natexlab{b}})}]{kim-klinger-2018-feels}
Evgeny Kim and Roman Klinger. 2018{\natexlab{b}}.
\newblock \href {https://aclanthology.org/C18-1114} {Who feels what and why?
  annotation of a literature corpus with semantic roles of emotions}.
\newblock In \emph{Proceedings of the 27th International Conference on
  Computational Linguistics}, pages 1345--1359, Santa Fe, New Mexico, USA.
  Association for Computational Linguistics.

\bibitem[{Liu et~al.(2019)Liu, Ott, Goyal, Du, Joshi, Chen, Levy, Lewis,
  Zettlemoyer, and Stoyanov}]{Liu2019Roberta}
Yinhan Liu, Myle Ott, Naman Goyal, Jingfei Du, Mandar Joshi, Danqi Chen, Omer
  Levy, Mike Lewis, Luke Zettlemoyer, and Veselin Stoyanov. 2019.
\newblock \href {https://arxiv.org/abs/1907.11692} {{RoBERTa}: A robustly
  optimized {BERT} pretraining approach}.
\newblock arXiv:1907.11692.

\bibitem[{Mohammad and
  Bravo-Marquez(2017)}]{mohammad-bravo-marquez-2017-emotion}
Saif Mohammad and Felipe Bravo-Marquez. 2017.
\newblock \href {https://doi.org/10.18653/v1/S17-1007} {Emotion intensities in
  tweets}.
\newblock In \emph{Proceedings of the 6th Joint Conference on Lexical and
  Computational Semantics (*{SEM} 2017)}, pages 65--77, Vancouver, Canada.
  Association for Computational Linguistics.

\bibitem[{Mohammad et~al.(2014)Mohammad, Zhu, and
  Martin}]{mohammad-etal-2014-semantic}
Saif Mohammad, Xiaodan Zhu, and Joel Martin. 2014.
\newblock \href {https://doi.org/10.3115/v1/W14-2607} {Semantic role labeling
  of emotions in tweets}.
\newblock In \emph{Proceedings of the 5th Workshop on Computational Approaches
  to Subjectivity, Sentiment and Social Media Analysis}, pages 32--41,
  Baltimore, Maryland. Association for Computational Linguistics.

\bibitem[{Plutchik(2001)}]{Plutchik2001}
Robert Plutchik. 2001.
\newblock \href {https://www.jstor.org/stable/27857503} {The nature of
  emotions}.
\newblock \emph{American Scientist}, 89(4):344--350.

\bibitem[{Pontiki et~al.(2014)Pontiki, Galanis, Pavlopoulos, Papageorgiou,
  Androutsopoulos, and Manandhar}]{pontiki-etal-2014-semeval}
Maria Pontiki, Dimitris Galanis, John Pavlopoulos, Harris Papageorgiou, Ion
  Androutsopoulos, and Suresh Manandhar. 2014.
\newblock \href {https://doi.org/10.3115/v1/S14-2004} {{S}em{E}val-2014 task 4:
  Aspect based sentiment analysis}.
\newblock In \emph{Proceedings of the 8th International Workshop on Semantic
  Evaluation ({S}em{E}val 2014)}, pages 27--35, Dublin, Ireland. Association
  for Computational Linguistics.

\bibitem[{Russell and Mehrabian(1977)}]{russell1977evidence}
James~A Russell and Albert Mehrabian. 1977.
\newblock \href {https://doi.org/10.1016/0092-6566(77)90037-X} {Evidence for a
  three-factor theory of emotions}.
\newblock \emph{Journal of research in Personality}, 11(3):273--294.

\bibitem[{Scarantino(2016)}]{Scarantino2016}
Andrea Scarantino. 2016.
\newblock The philosophy of emotions and its impact on affective science.
\newblock In \emph{Handbook of emotions}, chapter~4, pages 3--48. Guilford
  Press New York, NY.

\bibitem[{Scherer et~al.(2001)Scherer, Schorr, and Johnstone}]{Scherer2001a}
Klaus~R Scherer, A~Schorr, and T~Johnstone. 2001.
\newblock \emph{Appraisal considered as a process of multi-level sequential
  checking}, volume~92.
\newblock Oxford University Press.

\bibitem[{Smith and Ellsworth(1985)}]{Smith1985}
Craig~A Smith and Phoebe~C Ellsworth. 1985.
\newblock \href {https://doi.org/10.1037/0022-3514.48.4.813} {Patterns of
  cognitive appraisal in emotion.}
\newblock \emph{Journal of personality and social psychology}, 48(4):186--209.

\bibitem[{Stranisci et~al.(2022)Stranisci, Frenda, Ceccaldi, Basile, Damiano,
  and Patti}]{stranisci-etal-2022-appreddit}
Marco~Antonio Stranisci, Simona Frenda, Eleonora Ceccaldi, Valerio Basile,
  Rossana Damiano, and Viviana Patti. 2022.
\newblock \href {https://aclanthology.org/2022.lrec-1.406} {{APPR}eddit: a
  corpus of {R}eddit posts annotated for appraisal}.
\newblock In \emph{Proceedings of the Thirteenth Language Resources and
  Evaluation Conference}, pages 3809--3818, Marseille, France. European
  Language Resources Association.

\bibitem[{Troiano et~al.(2022)Troiano, Oberl\"ander, Wegge, and
  Klinger}]{Troiano2022-xenvent}
Enrica Troiano, Laura Oberl\"ander, Maximilian Wegge, and Roman Klinger. 2022.
\newblock \href {https://arxiv.org/abs/2203.10909} {{x-enVENT}: A corpus of
  event descriptions with experiencer-specific emotion and appraisal
  annotations}.
\newblock In \emph{Proceedings of The 13th Language Resources and Evaluation
  Conference}, Marseille, France. European Language Resources Association.

\bibitem[{Troiano et~al.(2023)Troiano, Oberländer, and Klinger}]{Troiano2023}
Enrica Troiano, Laura Oberländer, and Roman Klinger. 2023.
\newblock \href {https://doi.org/10.1162/coli_a_00461} {{Dimensional Modeling
  of Emotions in Text with Appraisal Theories: Corpus Creation, Annotation
  Reliability, and Prediction}}.
\newblock \emph{Computational Linguistics}, 49(1):1--72.

\bibitem[{Troiano et~al.(2019)Troiano, Pad{\'o}, and
  Klinger}]{Troiano2019-enisear}
Enrica Troiano, Sebastian Pad{\'o}, and Roman Klinger. 2019.
\newblock \href {https://doi.org/10.18653/v1/P19-1391} {Crowdsourcing and
  validating event-focused emotion corpora for {G}erman and {E}nglish}.
\newblock In \emph{Proceedings of the 57th Annual Meeting of the Association
  for Computational Linguistics}, pages 4005--4011, Florence, Italy.
  Association for Computational Linguistics.

\bibitem[{Wegge et~al.(2022)Wegge, Troiano, Oberlaender, and
  Klinger}]{wegge-etal-2022-experiencer}
Maximilian Wegge, Enrica Troiano, Laura Ana~Maria Oberlaender, and Roman
  Klinger. 2022.
\newblock \href {https://aclanthology.org/2022.nlpcss-1.3}
  {Experiencer-specific emotion and appraisal prediction}.
\newblock In \emph{Proceedings of the Fifth Workshop on Natural Language
  Processing and Computational Social Science (NLP+CSS)}, pages 25--32, Abu
  Dhabi, UAE. Association for Computational Linguistics.

\bibitem[{Wolf et~al.(2020)Wolf, Debut, Sanh, Chaumond, Delangue, Moi, Cistac,
  Rault, Louf, Funtowicz, Davison, Shleifer, von Platen, Ma, Jernite, Plu, Xu,
  Le~Scao, Gugger, Drame, Lhoest, and Rush}]{Wolf2020}
Thomas Wolf, Lysandre Debut, Victor Sanh, Julien Chaumond, Clement Delangue,
  Anthony Moi, Pierric Cistac, Tim Rault, Remi Louf, Morgan Funtowicz, Joe
  Davison, Sam Shleifer, Patrick von Platen, Clara Ma, Yacine Jernite, Julien
  Plu, Canwen Xu, Teven Le~Scao, Sylvain Gugger, Mariama Drame, Quentin Lhoest,
  and Alexander Rush. 2020.
\newblock \href {https://doi.org/10.18653/v1/2020.emnlp-demos.6} {Transformers:
  State-of-the-art natural language processing}.
\newblock In \emph{Proceedings of the 2020 Conference on Empirical Methods in
  Natural Language Processing: System Demonstrations}, pages 38--45, Online.
  Association for Computational Linguistics.

\bibitem[{Yuan et~al.(2020)Yuan, Fan, Bao, and Xu}]{yuan-etal-2020-emotion}
Chaofa Yuan, Chuang Fan, Jianzhu Bao, and Ruifeng Xu. 2020.
\newblock \href {https://doi.org/10.18653/v1/2020.emnlp-main.289}
  {Emotion-cause pair extraction as sequence labeling based on a novel tagging
  scheme}.
\newblock In \emph{Proceedings of the 2020 Conference on Empirical Methods in
  Natural Language Processing (EMNLP)}, pages 3568--3573, Online. Association
  for Computational Linguistics.

\bibitem[{Zhou et~al.(2016)Zhou, Shi, Tian, Qi, Li, Hao, and Xu}]{Zhou2016}
Peng Zhou, Wei Shi, Jun Tian, Zhenyu Qi, Bingchen Li, Hongwei Hao, and Bo~Xu.
  2016.
\newblock \href {https://doi.org/10.18653/v1/P16-2034} {Attention-based
  bidirectional long short-term memory networks for relation classification}.
\newblock In \emph{Proceedings of the 54th Annual Meeting of the Association
  for Computational Linguistics (Volume 2: Short Papers)}, pages 207--212,
  Berlin, Germany. Association for Computational Linguistics.

\end{thebibliography}
\clearpage

\appendix
\section{Distributions Emotion Spans and False Negatives}
\newcommand{\n}{\phantom{0}}
\begin{table}[h!]
  \centering\small
  \setlength{\tabcolsep}{12pt}
  \begin{tabular}{l rr rr}
    \toprule
    & \multicolumn{2}{c}{Writer} & \multicolumn{2}{c}{Non-Writer}\\
    \cmidrule(r){2-3}
    \cmidrule(r){4-5}
    Emotion 
    & \% & \# & \% & \# \\
    \cmidrule(r){1-1}\cmidrule(rl){2-3}\cmidrule(rl){4-5}
    anger & .61 & 204 & .39 & 132\\
    disgust & .76 & 66 & .24 & 21\\
    fear & .61 & 135 & .39 & 85\\
    joy & .45 & 118 & .55 & 147\\
    no emotion & .16 & 43 & .84 & 226\\
    other & .50 & 59 & .50 & 58\\
    sadness & .59 & 249 & .41 & 174\\
    shame & .64 & 209 & .36 & 116\\
    \bottomrule
  \end{tabular}
  \caption{Frequency (absolute and relative) of writer and non-writer spans annotated with a given emotion.}
  \label{tab:emo_freq}
\end{table}

\begin{table}[h!]
  \centering\small
  \setlength{\tabcolsep}{12pt}
  \begin{tabular}{l ccc}
    \toprule
    & total & \multicolumn{2}{c}{due to non-recogn.\ span} \\
    \cmidrule(rl){2-2}\cmidrule(l){3-4}
    Emotion & \#    & \# & \% \\
    \cmidrule(r){1-1}\cmidrule(rl){2-2}\cmidrule(rl){3-3}\cmidrule(rl){4-4}
    anger & 28 & \n7 & .25\\
    disgust & \n7 & \n5 & .71\\
    fear & 13 & \n4 &.31\\
    joy & 12 & \n7 & .58\\
    no emotion & 23 & 14 & .61\\
    other & 19 & \n4 & .21\\
    sadness & 21 & \n9 & .43\\
    shame & 21 & 10 & .48\\
    \bottomrule
  \end{tabular}
  \caption{Number of false negative emotion predictions (relative and absolute) that were introduced
    due to the experiencer predictor not recognizing the span.}
  \label{tab:emo_fn}
\end{table}

\newpage

\section{Distributions Appraisal Spans and False Negatives}

\begin{table}[h!]
  \centering\small
  \setlength{\tabcolsep}{11pt}
  \begin{tabular}{l rr rr}
    \toprule
    & \multicolumn{2}{c}{Writer} & \multicolumn{2}{c}{Non-Writer}\\
    \cmidrule(r){2-3}\cmidrule(r){4-5}
    Appraisal & \% & \# & \% & \# \\
    \cmidrule(r){1-1}\cmidrule(rl){2-3}\cmidrule(rl){4-5}
    suddenness & .62 & 333 & .38 & 202\\
    familiarity & .9 & 3 & .91 & 30\\
    pleasantness & .53 & 99 & .47 & 87\\
    understand & .58 & 642 & .42 & 460\\
    goal relev. & .47 & 40 & .53 & 45\\
    self resp. & .47 & 244 & .53 & 273\\
    other resp. & .50 & 256 & .50 & 251\\
    sit.\ resp. & .70 & 140 & .30 & 59\\
    effort & .33 & 25 & .67 & 51\\
    exert & .38 & 3 & .62 & 5\\
    attend & .44 & 18 & .56 & 23\\
    consider & .54 & 140 & .46 & 119\\
    outcome prob. & .54 & 211 & .46 & 177\\
    expect.\ discrep. & .60 & 380 & .40 & 252\\
    goal conduc. & .44 & 76 & .56 & 96\\
    urgency & .40 & 10 & .60 & 15\\
    self control & .39 & 136 & .61 & 217\\
    other control & .50 & 199 & .50 & 203\\
    sit.\ control & .67 & 135 & .33 & 67\\
    adj.\ check & .33 & 145 & .67 & 301\\
    int.\ check & .39 & 26 & .61 & 41\\
    ext.\ check & .21 & 9 & .79 & 34\\	
    \bottomrule
  \end{tabular}
  \caption{Frequency (absolute and relative) of writer-/non-writer spans annotated with a given appraisal class.}
  \label{tab:appr_freq}
\end{table}
\begin{table}[h!]
  \centering\small
  \setlength{\tabcolsep}{6pt}
  \begin{tabular}{l r r r}
    \toprule
    & total & \multicolumn{2}{c}{due to non-recogn.\ span} \\
    \cmidrule(rl){2-2}\cmidrule(l){3-4}
    Appraisal & \#    & \# & \% \\
    \cmidrule(r){1-1}\cmidrule(rl){2-2}\cmidrule(rl){3-3}\cmidrule(rl){4-4}
    suddenness & 30 & 11 & .37 \\
    familiarity & 3 & 1 & .33 \\
    pleasantness & 5 & 3 & .60 \\
    understand & 28 & 28 & 1\phantom{.00} \\
    goal relev. & 7 & 2 & .29 \\
    self resp. & 25 & 23 & .92 \\
    other resp. & 28 & 13 & .46 \\
    sit.\ resp. & 6 & 3 & .50 \\
    effort & 6 & 3 & .50 \\
    exert & 2 & 1 & .50 \\
    attend & 5 & 2 & .40 \\
    consider & 15 & 5 & .33 \\
    outcome prob. & 20 & 13 & .65 \\
    expect. discrep. & 41 & 16 & .39 \\
    goal conduc. & 9 & 4 & .44 \\
    urgency & 3 & 1 & .33 \\
    self control  & 23 & 19 & .83 \\
    other control & 33 & 12 & .36 \\
    sit.\ control & 6 & 3 & .50 \\
    adj.\ check & 36 & 20 & .56 \\
    int.\ check & 7 & 2 & .29 \\
    ext.\ check & 6 & 4 & .67 \\
    \bottomrule
  \end{tabular}
    \caption{Number of false negative appraisal predictions (relative and absolute) that were introduced
    due to the experiencer predictor not recognizing the span.}
  \label{tab:appr_fn}
\end{table}
\end{document}